\documentclass[11pt]{article}
\usepackage{geometry}
\usepackage{coling2020}
\usepackage{times}
\usepackage{url}

\usepackage{latexsym}
\usepackage{microtype}
\usepackage{graphicx}
\usepackage{subcaption}
\usepackage{amsmath}
\usepackage{hhline}
\usepackage{array, caption, floatrow, tabularx, makecell, booktabs}%
\usepackage{subcaption}
\usepackage{adjustbox}
\usepackage{floatrow}
\newfloatcommand{capbtabbox}{table}[][\FBwidth]
\colingfinalcopy 

\title{kk2018 at SemEval-2020 Task 9: Adversarial Training for Code-Mixing Sentiment Classification}

\author{Jiaxiang Liu, Xuyi Chen, Shikun Feng, Shuohuan Wang, \\
\textbf{Xuan Ouyang, Yu Sun, Zhengjie Huang, Weiyue Su} \\ 
  Baidu Inc., China \\
  {\tt \{liujiaxiang, chenxuyi, fengshikun01, wangshuohuan\}@baidu.com} \\
  {\tt \{ouyangxuan, sunyu02, huangzhengjie, suweiyue\}@baidu.com} \\}

\date{}

\begin{document}
\maketitle
\begin{abstract}
  Code switching is a linguistic phenomenon that may occur within a multilingual setting where speakers share more than one language. With the increasing communication between groups with different languages, this phenomenon is more and more popular. However, there are little research and data in this area, especially in code-mixing sentiment classification. In this work, the domain transfer learning from state-of-the-art uni-language model ERNIE is tested on the code-mixing dataset, and surprisingly, a strong baseline is achieved. Furthermore, the adversarial training with a multi-lingual model is used to achieve 1st place of SemEval-2020 Task 9 Hindi-English sentiment classification competition.
\end{abstract}

\section{Introduction}
\blfootnote{
    %
    %
   
    %
    %
    %
    %
     \hspace{-0.65cm}  
     This work is licensed under a Creative Commons 
     Attribution 4.0 International License.
     License details:
     \url{http://creativecommons.org/licenses/by/4.0/}.
}

In today's society, the use of social media has become a necessary daily activity. As a result, a large amount of text content has been produced. Among those corpora, there are many code-mixed texts. Code-mixing is a common phenomenon in societies in which two or more languages are used. It is the change of one language to another within the same utterance or in the same oral/written text \cite{ho2007code}. Code-mixing poses several unseen difficulties in NLP tasks on lexical/syntax/semantic levels.

In recent years, a lot of models based on Transformer \cite{vaswani2017attention} have been proposed such as BERT \cite{devlin-etal-2019-bert}, RoBERTa \cite{liu2019roberta} and ALBERT \cite{lan2019albert}. Those models take advantage of both available large corpora and computational power and have outperformed the traditional techniques in various sentiment analysis tasks \cite{kim-2014-convolutional,hochreiter1997long}. Meanwhile, many multi-lingual models are proposed, e.g. \cite{ma1999inhibition,conneau2019unsupervised}. \cite{10.1145/3342827.3342850} have tested the multilingual BERT with Hierarchical Attentive Network, \newcite{shakeel2020adapting} have tested BERT for code-mixing informal short text classification. 

However, the code-mixed corpus is small. Large models like BERT \cite{devlin-etal-2019-bert} and ERNIE \cite{sun2019ernie} have millions of parameters, and are prone to overfit. To tackle this problem, adversarial training can be used as a regularizer in the training phase \cite{szegedy2013intriguing,goodfellow2014explaining}. Adversarial examples are a way of fooling a neural network to behave incorrectly \cite{szegedy2013intriguing}. They
are created by applying a small perturbation to the original
inputs. The original algorithm was designed for use in the image field, and the perturbation added to the image was deliberately designed to be invisible to the naked eye, but causing neural networks to output a completely different response from the true one. 
By introducing those adversarial examples, which the neural nets make mistakes on, to the network during training, the performance can be improved. This is called “adversarial training” which
acts as a regularizer to help the network generalize better
\cite{goodfellow2014explaining}. For natural language, it is not feasible to produce perturbed examples due to the nature of the
text. As a solution, \newcite{miyato2016adversarial}
apply this perturbation to the embedding space and design a new loss for semi-supervised adversarial training.

Our contributions are twofold. First, the domain transfer learning method is applied for code-mixing classification, which surprisingly shows promising efficacy. And furthermore, the adversarial training is applied during the training process with a multilingual model. Finally, this method has achieved the state-of-the-art score and rank 1st in SemEval-2020 Task 9 for code-mixing Hindi-English sentiment classification.

\section{Related Work}

In this section, the existing research on multi-lingual sentiment classification and adversarial learning is discussed, which are closely related to this work.


\subsection{Multilingual Sentiment Classification}
Most of the state-of-the-art multilingual models are based on Transformer \cite{vaswani2017attention}. In mBert \cite{devlin-etal-2019-bert}, a 12 layers model trained on a multilingual corpus including 104 languages is proposed.
\cite{conneau2019cross} has proposed a new learning target to leverage both unsupervised and supervised corpus on monolingual and cross-lingual corpus respectively to train the cross-lingual language model XLM. \cite{conneau2019unsupervised} upgrades the XLM to XLM-R with more training data, which significantly outperforms mBERT and XLM in many datasets including a sentiment dataset SST-2  \cite{socher2013recursive} and a cross-lingual entailment classification dataset MNLI \cite{williams2017broad}.

\subsection{Adversarial Training}
Adversarial examples were first found in \cite{szegedy2013intriguing} where a malicious example called adversarial example can be designed while it is not perceptible for human eyes. Both \cite{szegedy2013intriguing} and \cite{goodfellow2014explaining} found that by putting the adversarial example back into training, the model will be regularized and can provide an additional regularization benefit beyond that provided by using dropout. Since those methods are proposed and tested in image classification, \cite{miyato2016adversarial} designed a semi-supervised loss based on adversarial training for sentence classification and shows promising results on text classification results. Therefore, in this work, the impact of applying adversarial training with the state-of-the-art multilingual language model is studied.

\section{Model}

In this section, we describe the two models that were used in this competition are proposed. The first one is ERNIE proposed by \cite{sun2019ernie} which uses the paradigm of continuous learning to pretrain language models which are adapted to sentiment classification using domain transfer learning. The other one is a multi-lingual model XLM-R and we have further trained it with adversarial training.

\subsection{Backbone Model}

The backbone models ERNIE \cite{sun2019ernie} and XLM-R \cite{conneau2019unsupervised} used in this tasks are all based on Transformer. For this code-mixing sentiment classification task, the input sentence will be  organized as
$$
[CLS], w_1, w_2, ..., w_n, [SEP]
$$
where the $[CLS]$ token is inserted in the beginning place of the sequence which is used as an indicator of the whole sentence, and specifically it is used to perform sentiment classification. The $[SEP]$ is a token to
separate a sequence from the subsequent one and is an indicator of the end of a sentence. $w_i$ is token of the sequence. 
After they go through the backbone
model, for each item of the sequence, a vector representation
of the size $H$, size of specified hidden layers, is computed.
The representation of $[CLS]$ is applied with a fully connected layer to classify the three sentiment labels including negative, neutral,  and positive.

\subsection{Training with Mono-lingual Model}

The ERNIE model is trained with a continuous learning method. The training tasks of ERNIE include different granularity, such as the lexical level task of capital letter predictions and masked language model for entity, syntax level task of predicting document order, and relation, e.g. entailment relation prediction.

This model is trained with monolingual corpus, including Wikipedia, BookCorpus, and Reddit. The off-the-shell model is used directly for finetuning on the code-mixing corpus.

After the code-mixing sentence goes through ERNIE, 
the representation of the token $[CLS]$ is used to classify the sentiment. Softmax is used to calculate the probability distribution of each class. The cross entropy is used to calculate the loss:

 $$ L_{\mathrm{ce}} = -y \log p(y|x;\theta)$$
where, $p(y|x)$ is the probability for target label $y$.

\begin{figure}
\begin{floatrow}
\ffigbox{  \includegraphics[width=0.45\textwidth]{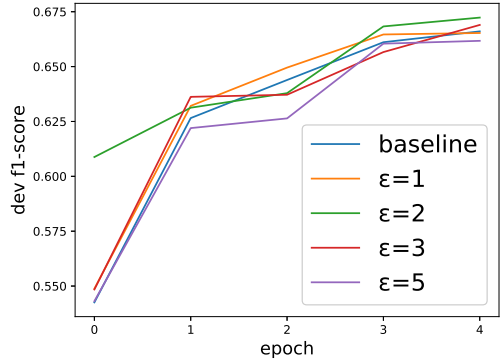}
}
{%

}\label{fig1}

\end{floatrow}
\caption{ The effect of the hyper-parameter $\epsilon$ for adversarial training }
\end{figure}

\subsection{Adversarial Training Multilingual Model}

XLM-R \cite{conneau2019unsupervised} is used as the multilingual backbone model for this task. XLM-R is trained with 100 languages with 2.5 TB corpus. The training objective is the same as RoBERTa \cite{liu2019roberta} which is a masked language model, that is,  the input is a corrupted sentence and the goal is to recover those corrupted tokens.

Adversarial examples are inputs to machine learning models that an attacker has intentionally designed to cause the model to make a mistake \cite{goodfellow2014explaining}.
According to the access to the model parameters, it can be divided into two types of attack algorithms: white box attack and black box attack. The former adversarial one has full access to the model parameters (e.g. embedding, linear mapping weights), and the latter uses only on the input and output, e.g. token id, the probability distribution of model prediction.
Considering that the goal of adversarial training in our experiment is to regularize the training process, the white-box method allows it to access the embeddings. 

To create adversarial examples, the formula proposed by
\cite{miyato2016adversarial} is used, where the perturbation is created
using the gradient of the loss function. Assume the input embedding $x$ and the model
parameters $\theta$ is given, to find the adversarial examples the
following problem should be solved:

$$
\boldsymbol{r}_{\operatorname{adv}}=\epsilon g /\|g\|_{2} \quad \mathrm{where} \quad g=\nabla_{x+d} \mathrm{KL}[p(\cdot | \boldsymbol{x} ; \hat{\boldsymbol{\theta}}) \| p(\cdot | \boldsymbol{x}+\boldsymbol{d} ; \hat{\boldsymbol{\theta}})]
$$
where $\mathrm{KL}[p\|q]$ denotes the KL divergence between distributions $p$ and $q$, $r$ denotes the perturbation on the input and $\hat{\theta}$ is a
constant copy of $\theta$ in order not to allow the gradients to
propagate in the process of constructing the artificial examples. Solving the above problem means that
we are searching for the worst perturbation while trying to
maximize the divergence of the outputs between benign input $x$ and adversarial input $x+d$ of the model, where $d$ is a small random vector. Therefore, the perturbations are injected to the input embeddings to create new
adversarial sentences in the embedding space. $\epsilon$ is the size of the perturbation. To find values
that outperform the original results, an ablation study was carried out on four values for $\epsilon$ whose results are
presented in Figure 1 and discussed in Section Model Robust Analysis. After the

\vspace{0.5cm}
\capbtabbox{
\begin{tabular}{|l|l|l|l|}
\hline
       &  Precision           & Recall              & F1                   \\ \hline \hline
Positive     &    0.8426           &  0.7600        &  0.7992            \\ \hline
Negative     &    0.7850           &  0.7544        &  0.7694           \\ \hline
Neutral      &    0.6521           &  0.7309        &  0.6892               \\ \hline
Macro Average      &    -                &  -             &  0.7526               \\ \hline
\end{tabular}
}{
\caption{ Results of our test set submissions
 }
}\label{tab1bb}
adversarial examples go through the network, their loss is
calculated as follows:
$$ 
L_{\operatorname{adv}}(\boldsymbol{\theta})=\mathrm{KL}\left[p\left(\cdot | \boldsymbol{x} ; \hat{\boldsymbol{\theta}}\right)|| p\left(\cdot | \boldsymbol{x}+\boldsymbol{r}_{\mathrm{ad} \mathbf{v}} ; \boldsymbol{\theta}\right)\right]
$$
Effectively, the total loss is calculated by 
$$ 
L = L_{\mathrm{ce}} + \alpha L_{\mathrm{adv}} 
$$  as the adversarial term favors label smoothness in the embedding neighborhood, and $\alpha$ is a hyperparameter that controls the trade-off between standard errors and robust errors.

\section{Experiments}

\textbf{Implementation details}
There are  14000/3000/3000 examples in the train/dev/test dataset respectively, for more detailed dataset information please refer to the work of \cite{patwa2020sentimix}.
To train a better model, data cleaning strategies have been applied, e.g. the URLs, hashtags, usernames are all removed. When processing the data, to increase the data diversity, we tried both retaining and discarding case information. To fully make use of the data provided by the competition, the Spanish-English code-mixing dataset is combined with Hindi-English dataset, however, it shows a negative effect with training those two datasets together. 
 We performed all our experiments on a GPU (Nvidia Tesla V100) with 32 GB of memory. Except for the code specific to our model, we adapted
the codebase utilized by XLM-R. To carry out the ablation study of adversarial training, batches of 32 were specified. For optimization, the
Adam optimizer \cite{adam} with a learning rate of $3e-5$ was used. Also, we have applied the linear warm strategy for the learning rate. We set $\alpha=1$ for the adversarial loss. The experiments are conducted for 5 epochs and we pick the best model using the dev set.

\textbf{Improving Generalization}
\label{analysis}
As the adversarial training can be a regularizer to model, it is expected to have a more robust generalization on both development and test set. From Figure 1  it can be seen that the hyperparameter $\epsilon$ is important. When $\epsilon = 2$, we get the best F1 score for the development dataset, and furthermore, the scores for this parameter are better than the baseline most of the time (green line in Figure 1). From Table 2, there is a trend with decreasing development set score but increasing score in test set score, which shows the overfitting is easing. From this, we can see the increasing generalization and robustness of the trained model.

\textbf{Domain Transfer Learning Discussion}. 
From Table 2, it can be seen that ERNIE is tested as the baseline. ERNIE is proposed as a uni-language pretrained model trained for English. We have applied the model finetuned on the Hindi-English code-mixed dataset directly as domain transfer learning, and surprisingly this model provided a strong baseline (0.6725 test score) for this task. We speculate that this result comes first because of the structural similarity between the two languages as they both belong to Indo-European languages, which is important and fully discussed in the in \cite{DBLP:journals/corr/abs-1906-01502}, and second, thanks to the BPE tokenizer \cite{sennrich-etal-2016-neural}, tokens can be processed at sub word level, thus reducing the proportion of unknown words. As a result, the ERNIE shows a strong domain transfer learning ability.

\vspace{0.5cm}
\capbtabbox{
\begin{tabular}{|l|l|l|}
\hline
Models           &  Dev set score     & Test set score                                      \\ \hline \hline
ERNIE 2.0        &    0.6571            &  0.6725                                             \\ \hline \hline
XLM-R            &    0.6660            &  0.7177                                               \\ \hline
+$\epsilon$=1      &    0.6653            &  0.7305                                               \\ \hline
+$\epsilon$=2      &    0.6723            &  0.7320                                               \\ \hline
+$\epsilon$=3      &    0.6689            &  0.7332                                               \\ \hline
+$\epsilon$=5      &    0.6674            &  0.7349                                                  \\ \hline
Ensemble         &    --                &  \textbf{0.7526}                                      \\ \hline
\end{tabular}
}
{
\caption{Score analysis for candidates of final ensemble }
}\label{tab1}

\textbf{Ensemble Results}
The training set is very small compared to the large scale of the pre-training corpus and model parameters for the backbones we have used. To take full advantage of the training dataset and increase training diversity, the strategy of constructing a training dataset is important. In practice, a 10-fold cross validation is executed. During each fold, we obtain results for the test dataset which provides a large number of candidates for the ensemble. 
Besides, we also used different backbone network and hyperparameter settings to increase the result diversity. The ensemble method we used was mainly based on the average prediction result, that is, the probability of each category was averaged across all the ensemble candidates, and finally the category with the highest probability among the three categories was selected as the prediction result. As a result, we have reached 0.75 F1-score and achieved the 1st place in competition. 

However, from Table 1, it can be seen the F1 score of the neutral class is considerably low, which is mainly caused by the low precision of the neutral class. We speculate that, despite using the adversarial training as a regularizer,  the model is still more inclined to learn specific keywords as emotional category indicators, resulting in low precision for neutral class. For the positive and negative class, the model performs better, but both the precision is higher than the recall score, which shows that the sentiment is not always obvious. We leave this research in future further work.

\section{Conclusion}
In this paper, we tested the performance of ERNIE for domain transfer learning and applied adversarial to code-mixed sentiment analysis with XLM-R. The experiments show that ERNIE provides a strong baseline in the cross-lingual setting. After replacing the backbone model with the multilingual model XLM-R, the F1-score of sentiment classification was significantly improved, further improved after the adversarial learning was applied. In particular, we have carried out experiments on hyperparameters in adversarial learning. Different parameters have a significant influence on the experimental results, but they all achieve the purpose of enhancing the generalization of the model. As a result, we have achieved the top rank for the competition of Hindi-English code-mixed sentiment classification and our user name is kk2018. As future work, other white-box adversarial examples as well as black-box ones will be utilized to compare adversarial training methods for various sentiment analysis tasks.

\newpage
\bibliographystyle{coling}
\bibliography{semeval2020}

\end{document}